# A Study of Image Pre-processing for Faster Object Recognition

Md Tanzil Shahriar
*Department of Computing Science*
*University of Alberta*
Email: mdtanzil@ualberta.ca

Huyue Li
*Department of Computing Science*
*University of Alberta*
Email: huyue@ualberta.ca

*Abstract*—Quality of image always plays a vital role in increasing object recognition or classification rate. A good quality image gives better recognition or classification rate than any unprocessed noisy images. It is more difficult to extract features from such unprocessed images which in-turn reduces object recognition or classification rate. To overcome problems occurred due to low quality image, typically pre-processing is done before extracting features from the image. Our project proposes an image pre-processing method, so that the performance of selected Machine Learning algorithms or Deep Learning algorithms increases in terms of increased accuracy or reduced the number of training images. In the later part, we compare the performance results by using our method with the previous used approaches.

## I. INTRODUCTION

Object recognition is a technology that detects objects of a class in digital images and videos. There are three main tasks of object recognition: Image classification, Object localization, Object detection. Image classification means predicting the class of objects in images. Object localization is identifying the location of objects in images and circle the objects out with bounding boxes. Object detection is a combination of Image classification and object localization. Although state-of-the-art deep learning methods can reach high accuracy in these tasks, they need to consume a lot of time to complete the model training. Our research aims to reduce the training time of object recognition models, starting with the task of object detection, we try to find a method that accelerates the training procedure while maintaining the initial accuracy.

We will focus on the processing of the dataset instead of modifying the existing Machine Learning-based Object Detection algorithms. We will choose any machine learning algorithm for example, basic SVM[9] to any deep learning algorithms like CNN. We will adjust the dataset that is used in the model training process without changing the algorithm. Typical neural network for object recognition can be selected as ideal algorithm, For example, Faster R-CNN [1] or YOLO [8]. After finalizing the algorithm, We will fine-tune the Real-Time Object Detection on PASCAL VOC 2007 [2] dataset to achieve faster convergence of the model. Our object is to provide a fine-tuned PASCAL VOC 2007 dataset, based on which the network's training time for a certain accuracy is shorten. Noticing that based on our progress, the fine-tuned dataset of PASCAL VOC 2007 might be in a certain class like a car, an animal, or face rather than full dataset. If time permits, we will expand the types of classes as many as possible.

## II. RELATED WORK

### A. Image Processing Methods

Fiala et al. used the Hough transformation for feature detection in Panoramic images [10]. These images have a 360° field-of-view. These images have advantages when dealing with computer vision problems such as navigation, object tracking, and world model creation. Previous research purposed Catadioptric imaging systems, that employing mirrors and lenses in the optical path, can capture a panoramic image. However, this device often fails when detecting straight lines. In this paper, A new method is proposed to locate horizontal line features. the existence and location of horizontal lines are found by mapping edge pixels to a new 2D parameter space for each mirror lobe. This method's name is Panoramic Hough transform. Panoramic Hough transform is derived from the Hough transform which identifies lines or curves by creating a separate parameter space and finding groupings in this parameter space. This new method transfers the Hough transform of the plane space to the panoramic picture. One can detect the shape feature of the object in the real world from a panoramic view. It has a good potential for detecting the distorted horizontal lines, with the robust performance achieved with the only estimated calibration of the image sensor. Besides that, its performance is promising even on images that have no preprocessing like denoise. This may imply that this method can be used in board conditions. The main idea of this paper is transforming the Hough transform into a certain field, in this case, the panoramic images. It is designed to solve the feature detection problem in the panoramic image. We may derive this thinking in our project where we intend to design a faster image preprocessing method for image segmentation. With the robust feature of the panoramic hough transform, we can easily implement this method for our project. For instance, Google street view images have the defect of panoramic distortion. If images from google street view are our target, we can implement the panoramic hough transform to them. In this way, we would get a clear street outline rather than a distorted version.

Basu et al. provided a novel image compress method called variable resolution specially designed for Teleconferencing to reduce the bandwidth required for transmission [11]. Data compression can be either lossless or lossy, and intuitively speaking, lossy methods tend to have better compression ratios. And VR transform is a lossy compression. This transform's idea comes from the animals' vision system in the dynamic world. This vision system focuses more on areas of greater interest. In this area (fovea), more detail is preserved while the outside area(periphery) requires less detail. At the time when the paper came out, there are no image compression applications that use the method. Besides, two advantages are shown in this method. The first one is the compression ratios are obtained by controlling the sampling. The other is that VR transforms process images using a look-up table, which will accelerate the compression procedure. From the algorithm's analysis, we know that video conferencing is an ideal field to implement the VR transform because of the requiring compression for video transmission across limited bandwidth channels. This transformation has two parameters, the expected savings, and alpha which controls the distortion at the edges of the image. When alpha is high, the fovea is sharply defined and the periphery is poorly defined, vice versa. But traditional VR transform has a defect that the compressed image is not rectangular. This problem is solved by using multiple scaling factors to make sure the transformed image can be mapped to a rectangle with full space utilization. Moreover, real images with multiple foveas are compressed using a weighted transformation, which avoids foveas misplacement. At last, the compression performances are evaluated based on three criteria: the compression ratio, compression time, and resulting image quality. The VR transform has promising performances in these criteria, especially in the compression time, where the method save 50 percent compression time in certain situation. Image compression is almost one solution that we will implement in our image data preprocessing project. Therefore, knowing the series of image compression method is one of our major topics. This paper provides us an example of image compression algorithms. The author takes VR-transform to image compression and evaluate it with the statistical method and compare it with other image compression methods. In this case, we get good intuition about the advancing level.

Firouzmanesh et al. purposed a novel way to compress 3d Motion capture data [12]. Motion capture is the process of recording movement and translating that movement onto a digital model. This data can consume a large amount of bandwidth. In this paper, The authors proposed a fast encoding and decoding technique for lossy compression of motion capture data, considering human perception. Experimental results show the algorithm is much faster than other comparable methods. The compression rate that they achieved is 25:1 with little impact on perceptual quality. Besides the advantage of the compression ratio, this algorithm is very efficient. It can be planted in real-time online 3D games that have a large amount of data flow. One reason they achieved such success is that

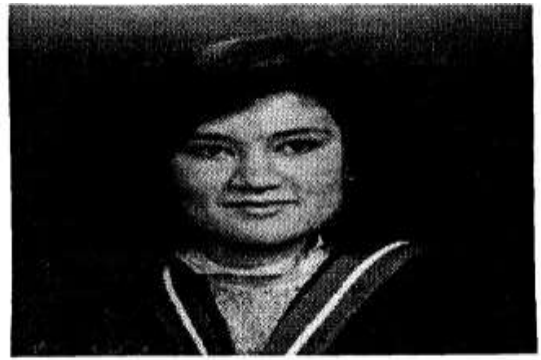
Figure 1: Original image.

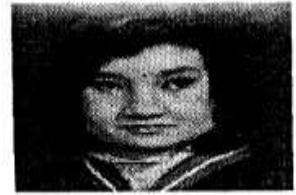
Figure 2: Compressed image; fovea near center.

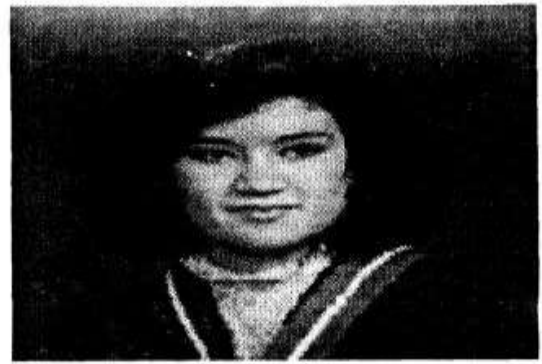
Figure 3: Uncompressed image

**Fig. 1: Variable Resolution transform**

They achieved perceptually guided compression of motion data by considering the human visual system when compressing 3d motions. Their method is derived from the Wavelet-Based Approach. The wavelet-based approach is applying a one-dimensional wavelet transform on each channel of motion data. Then they preserve a certain percentage of the wavelet coefficients while zero out the rest. After that, the quantization process follows to produce the compressed output stream. While achieving a high compress ratio, this compression is very time-consuming. What's worse, because of the high dependency of error metric on the joint hierarchy, giving more weight to the joints with high error rates does not yield good results. Considering the need for fast compression and availability for human perceptual factors of animation, the authors upgraded the traditional wavelet-based approach. Their idea is to select different numbers of coefficients for different channels of data based on the importance of each channel on the perceived quality of motion and based on the global properties of the object. One coefficient selection algorithm considered factor is the length of the bone connected to a joint. Intuitively speaking, larger bones have a greater effect on the perceptual quality of animation than smaller ones. After implementing this method, the main evaluation job is for real people. Comparing the proposed method and standard wavelet compression, the purposed method has a significant advantage. Besides, it also has good efficiency.

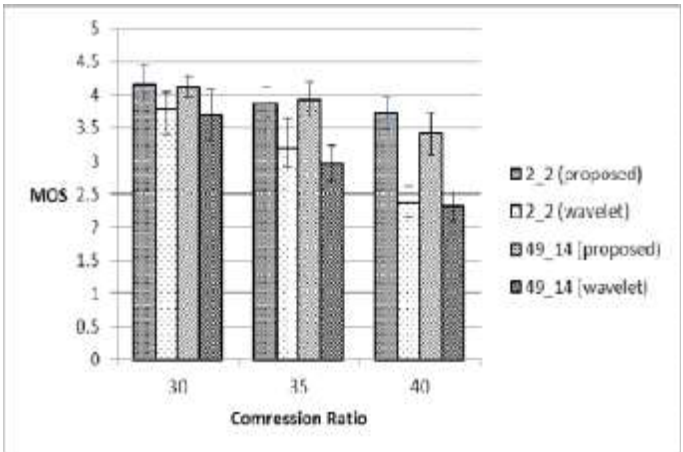

**Fig. 2: Efficiency comparison for proposed method**

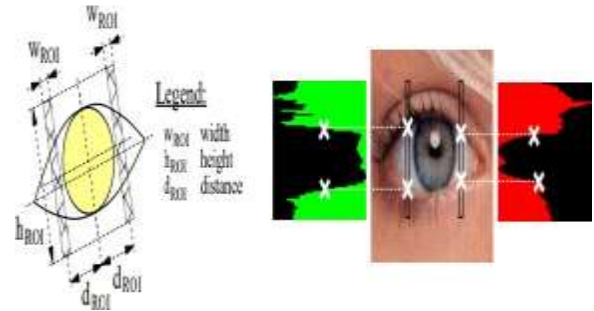

**Fig. 3: Eyes's location extraction parameters**

Bernogger, S et al. purposed an eye tracking method used in video to capture eyes' motion inspires us about the idea of POI tracking [13]. Nowadays, CG technology often extracts animation from real action from individuals. For now, we can extract the body's action easily. But as for facial detection, automatic facial feature detection and tracking along with facial expression analysis and synthesis still poses a big challenge. Eyes, as the most active organ in the human body, must be precisely detected in such work. In this paper, The authors purposed an algorithm for accurate localization and tracking of eye features. This algorithm is derived using the heuristic approach, from facial feature detection and tracking algorithms in other parts of the body. In this way, they developed the method especially for tracking and modeling of eye movements. Combined with processing like adjusting color images, Hough transformation, and deformable template matching, this algorithm gets a very accurate result. After extracting the features, another important component comes into being. It is synthesizing facial movements and expressing them at a remote site. Fortunately, the authors also provide a solution to that. They present an approach to synthesizing the eye movement by using the extracted eye features to compute the deformation of the eyes of the 3D model by mapping the model to the first frame of the face sequence then applying the extracted parameters to the 3d model. For the eye's feature extraction part, firstly, the Hough transform is implemented to detect the iris. Secondly, the deformable templates for eyelids are initialized. After we get the features' parameters, we map these parameters to the 3D model. Instead of using all the

Parameters. They use the most significant part of the eyes' feature to map the model. Finally, 'The eyes' movement is showed in the 3d model from the start frame sequence to the end. The experimental results show that by implementing the Hough transform. This feature extraction method is very robust against noise as well as edges that are not produced by the contour of the iris.

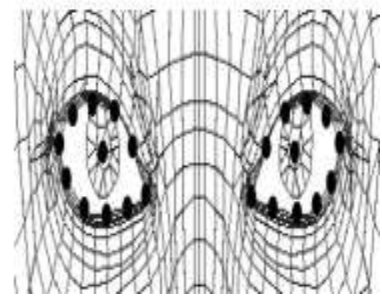

**Fig. 4: Eyes key points mapping**

Shen, Rui et al. propose a novel MEF method based on perceptual quality measures that exploit both contrast and color information [14]. This method is called the Quality of Experience (QoE) based MAP-HMGCRF fusion method. It generates images with better quality than existing ones for a variety of scenes. And here is another novelty about this method, human perceptual parameters are taken into consideration to evaluate pixel contributions more accurately. The following is a brief introduction. Given a source image sequence, the contributions from individual pixels to the fused image are perceptually tuned by perceived local contrast and color saturation. After this process, the perceived contrasts(luminance and color) are used in the MAP-HMGCRF model for pixel-level contribution evaluation. HMGCRF is a hierarchical architecture of MGCRF. It is used to efficiently estimate the pixel contributions/fusion weights(MAP) configuration on a lattice graph. And this method's result is promising from both object and subject view. The purposed method QBF beats other fusion methods in many criteria, like QAB/F and DRIVDP. Besides that, the research team even invites people to evaluate the outputs of the methods. In both views, The purposed image fusion method has a good performance. This paper's novelty is that it extended the mechanic image fusion research field to the human perception field. For example, when they were deciding parameters, they considered in which way the real human will evaluate the fusion image. From this step, they got the perceived local contrast and color saturation as the depending parameter. And their focus on the performance is not merely in the accuracy level, but also in time level. So that they derived HMGCRF from the slower MGCRF. This consideration may imply that the method has good applicability in the industry. From their research. We get to know what we should pay more attention to when designing a new algorithm. Besides that, we get the idea that the algorithm's criteria should not be limited. We may need to explore other research fields when designing our image preprocessing methods.

*B. Machine Learning Methods*

Taha, Bilal et al. managed to extract compressed 2d data from 3d models in order to decrease model training time [15].Compared to 2D photometric images, 3D data provides more information. However, The stored 3D data requires more storage compared to the 2D images. It is crucial to extract the 3D data's features while avoiding to be lossy. The focus of previous research on representing 3D data has been to carefully design hand-crafted methods of feature description. While automatically learned feature representations in terms of activations of a trained deep neural network have shown their superiority on many tasks using 2D RGB images, the authors purposed a novel method to deal with 3D data similarly. They proposed an approach to extend the application of deep learning solutions to 3D data given in the form of triangular meshes by exploiting the idea of establishing a map between the 3D mesh domain and the 2D image domain. Intuitively speaking, it maps the 3d data into a 2d image while preserving its features. There is a close work from other researchers, but the author's method is more computationally efficient compared to other works. Here is an overview of the purposed method. Firstly, they perform a down-sampling on the facial surface derived from the full 3D face model. This process is similar to the down-sampling process in 2D images. Then, To align the new 3D model and 2d texture, they designed an algorithm to map the texture image on the compressed mesh. In the end, they extract a group of local shape descriptors that contain most information from the original 3D model. These local descriptors can be computed efficiently and complement each other by encoding different shape information. After these transform, A novel scheme is then proposed to map the extracted geometric descriptors onto 2D textured images.

These images are called GAIs. Each image corresponds to a 3D model and its texture. Then GAIs are fused into three-channel images called FGAIs. At last, An CNN network is implemented to extract the features of FGAIs. Eventually, the output is the compressed 3D data. These data can be used in various of CNNs in a diverse area. The fig showed is an overview of the whole process.

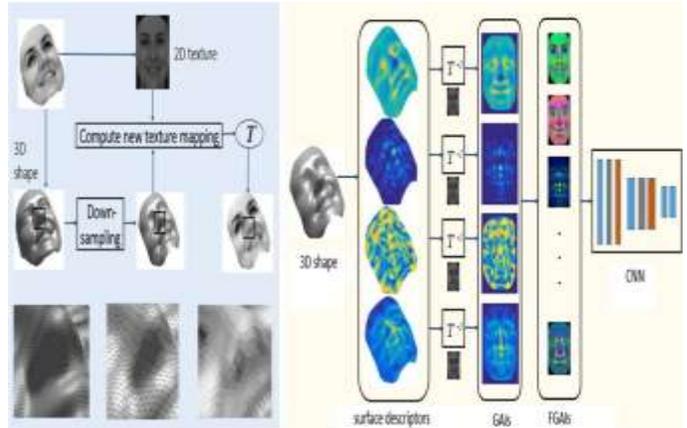

**Fig. 5: 3D Extract process**

According to the experiments tested on various classic CNN networks like VGG and AlexNet, We notice that computational efficiency is highly promoted while a slight accuracy percentage is influenced. In conclusion, this method shows a neat boost in performance when compared to competitive methods.

Kushal Sharma et al. found an novel way for Image compression [16]. Image compression is the process of optimizing image size without compromising much on the quality of the image. There are two main ways– lossless and lossy. The traditional image process deals with image compression problems in the field of spatial frequency. The theory is that areas in the image with low frequency contain less information compared to the high-frequency domain. Thus, we can compress one image by downsampling the low-frequency part. With the development of machine learning algorithms, the CNN based models are used to detect certain parts of images that need to be compressed. In this paper, the author proposes a lossy compression approach developed using a CNN based model. It is a combination of existing image classification CNN network–ResNet and a classical image compression paradigm Joint Photographic Experts Group(JPEG). Figure below showed their overall process. As for the network part. The reason they chose the Resnet model as the fundamental is that the model can be transferred as a Region of Interest extraction model with only a few configurations. Their modifications are developed on top of the Class Activation Mapping. By deleting the average pooling, flattening, and the final connected layers in the network, and replacing them with a 2D convolutional layer, 2-D depthwise convolution, and finally another convolutional layer, they can identify the regions of interests in the image. Eventually, they got a heatmap for each image showing the region of interest. After knowing the region of interest, they transferred the information to JPEG which consists of two parts - an encoder and a decoder. The encoder divides the image into blocks and applies DCT to obtain coefficients. The decoder performs the inverse operations when the image is displayed. They split the compression regions into three different areas, each of them has a different compress percentage. This is like a lookup table for JPEG to process the image. Their result showed that this model achieves compression ratios better than JPEG while maintaining a similar level of visual quality. But there are still a few flaws. One thing is that it needs a large amount of time to compress images of high resolutions. Another one is the minor degradation for images with no definite region of interest. We are looking forward to refining this method by trying a different network to achieve a better ROI extraction result.

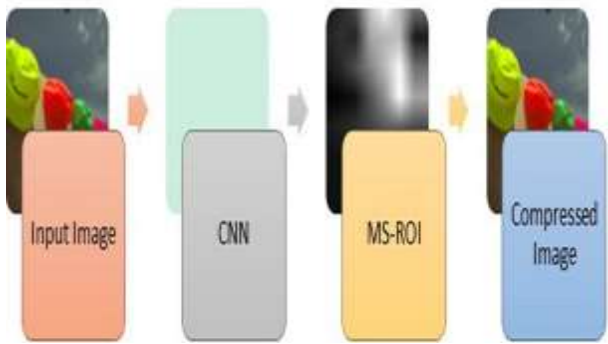

**Fig. 6: Image compression approach**

Graham-Knight, Brandon et al. trained a model that accurately segment the kidney CT images, which we can refer to. Kidney stone disease is the formation of calcifications within the kidney [17]. When kidney stones become large, obstruction in urine flow can occur, causing extreme pain, kidney damage, and even renal failure. About 12 percent of the world population is affected by kidney stones at some stage of their life time. And every year, the United States spends 5 billion dollars on kidney stone treatment. One of the most popular treatment is The PCNL, it involves the creation of a tract from the skin to the kidney through the flank using ultrasound and/or fluoroscopic guidance. During the treatment, CT imaging can aid urologists in determining the method to use. However, the method is not perfect because the CT images do not provide a conclusive picture of surgical outcomes. The urologists prefer the surgical orientation that CT image fails. To solve this problem, the authors purposed a model for kidney segmentation in CT scans, which is trained using the publicly-available KiTS19 dataset. This model is trained using nnU-Net which is derived from U-Net. U-Net is widely used in the field of medical image analysis such as MRI and CT scans. The training was done using 5-fold cross-validation, and the training data is from the 2019 Kidney Tumor Segmentation Challenge. As the figure shows, The blue line indicates training loss and the red line indicates validation loss. The green line represents the evaluation metric, its scale is on the right axis. We can see that the overall accuracy is above 0.8 and in some cases, it can exceed 0.9. Although the model produced excellent results for kidney segmentation, the model was unable to interpret the CT scan and produced no segmentation in some cases. This is because CT scans have a high variation. Despite the overall success, the model appears sensitive to changes in features in the datasets, with some segmentation masks working very well and others unable to correctly separate the kidney from the surrounding anatomy.

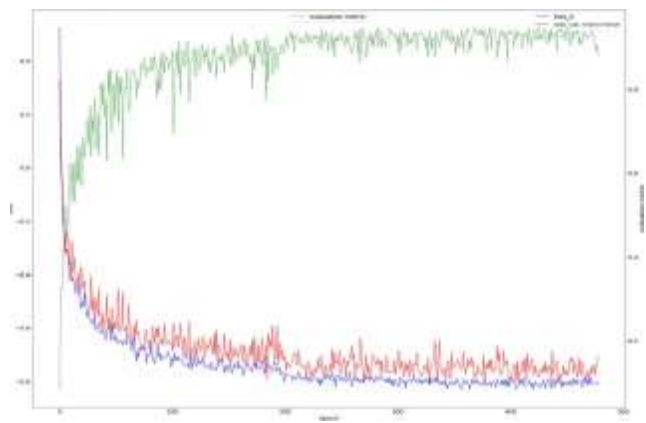

**Fig. 7: KIDNEY segmentation result**

Rossol, Nathaniel et al. provided a framework for Virtual reality rehabilitation [18]. It is an effective way to supplement patient rehabilitation. In this paper, the authors designed a flexible, low-cost rehabilitation system that uses virtual reality training and games to engage patients' rehabilitation on the use of powered wheelchairs. Besides that, a framework based on Bayesian networks for self-adjusting adaptive training in virtual rehabilitation environments is purposed. This method is proven effective according to the user's feedback. This framework has the following features: portable, indoor, and low-overhead. Unlike other training systems, this framework allows clinicians to design their training methods. In this way, They may find the rehabilitation process more appealing. And what's more promising is that the framework can measure patients' ability changes over time with Bayesian networks to better arrange the training process. Here is a brief introduction to how the system works. Firstly, this system is a software that can run on computers of systems. Its control system has an amazing moldability, which means that users can modify their control system based on their training part. And this system is built on a game engine named Open-Source Graphics Rendering Engine(OGRE). Secondly, it uses Bayesian networks to attempt to determine patient skill levels to create immersive and custom experience for individuals. This network scientifically evaluates an individual's performance on a certain task and feedback future's training intensity. From the result evaluation, Participants commonly have a better performance in real-world obstacle courses after training on the purposed wheelchair. The mean time advantage is 81.5 seconds in the trained group compared to 104.5 seconds in the untrained group. However, further investigation needs to be done. This is because of the high variance of individuals. Besides that, the training was done by individuals without any defects. Because of its price advantage and ease of use, I think this method has a bright prospect.

Comments:This paper was published more than 20 years ago, but it is still a valuable learning source. We are more focused on its method evaluation part. Firstly, it proved that strategy 1theoretically has better performance than strategy 2. But the outputs showed that the later one actually performs better.Then after detailed calculation, the author proved that it is because of the effect of image noise. This theory made the whole paper self-explanatory. Back to our project, it is likely that our method has good performance theoretically but not working well in real-world data. This paper provides us a possible way to continue our further research.

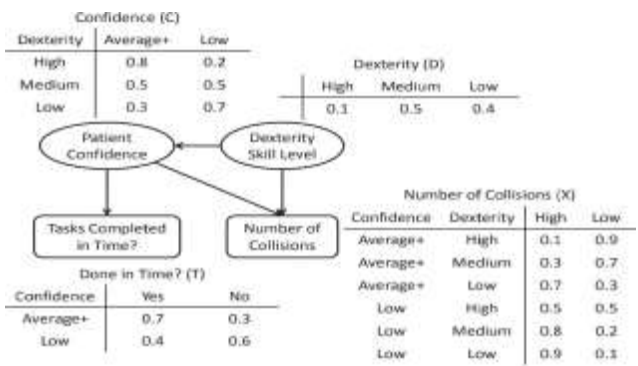

**Fig. 8: Bayesian network**

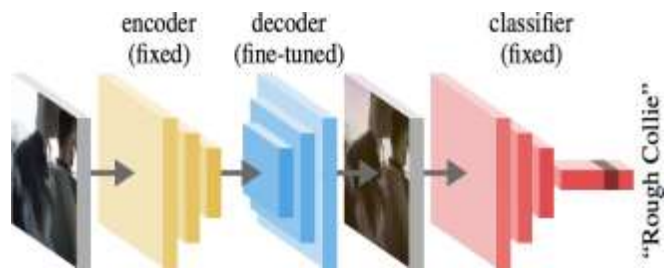

**Fig. 9: Architecture for Autoencoder**

Basu, Anup. purposed an automatic camera calibration methods to estimates the parameters of a lens and image sensor of an image [19]. It is often used in fields like correcting for lens distortion, measuring the size of an object in world units, or determining the location of the camera in the scene.At the time when this paper was written, people are suffering from the cameras' calibration problem–the algorithms for cam-era calibration is limited. There were two main method groups:linear methods that are easier to implement but incapable of lens distortion problem and non-linear methods that have good performances in complex situations but expensive to calculate.However, the author purposed a new dynamic solution for camera calibration that is derived from human vision. The initial observation is human's eyes can automatically calibrate when fixed on an object. And we get the sense because our eyes rotate when looking at something. By using the small pad and tilt movements, an active camera can estimate its center. The author provided two strategies. No.1 is to use three distinct image contours, or three different positions of the camera to estimate the error in the estimated lens center.Then these estimates can be used to estimate Focal length in pixels. No.2 Using a single image contour, estimate focal length in pixels first and use the length to estimate the error in the estimated lens center. By comparison, No.2 is more robust than No.1 even though No.1 is theoretically more precise. It is because of the noise that existed in the real world.

Christopher West et al. focused on diagnosing Parkinson's disease (PD) using four deep-learning based models that actually classified patients of PD based on the biomarkers that were found in structural magnetic resonance images [20]. Their 3D-Convolution-Neural-Network model demonstrates high efficacy in diagnosing Parkinson's disease, with an accuracy of 75% and 76% sensitivity. Also, they highlighted potential biomarkers for PD in cerebellum and occipital lobe.

They used T1-weighted sMRI data to classify patients and developed multiple CNN models for that purpose. They took the Parkinson's Progression Markers Initiative (PPMI) public dataset for these models. First they resized all the sMRI scans to the same dimension so that they perfectly fit for their model. They chose two different approaches to classify the patients for the deep-learning models. First one is based on three-dimensional analysis and on the other hand, the second one is strictly two-dimensional. All of the models were built using TensorFlow libraries.

All of the models performed better than the baseline class-distribution. The 2D models had very high sensitivity but less than satisfactory level. The best performing 3D model was the 3D-CNN, with 75% accuracy and balanced metrics in sensitivity, specificity and precision.

Leland Jansen et al. proposed an image feature representation technique which classifies different objects in 3D point clouds by using several 2D perspectives and then they used YOLO (v3) as object classification tool [21]. Proposed algorithm performed better in classifying pedestrians and vehicles in the Sydney Urban Objects data set, where their technique has achieved a classification accuracy of 98.42% and an f1 score of 0.9843.

They represented the 3D point clouds as a series of 2D rasterized images generated from different perspectives. Then the images were integrated into one single raster and was fed into YOLO. YOLO was used in this model for object detection. The proposed method consisted of three steps. First one is preprocessing and the other two are training the images and inference which were done using YOLO. They used the Sydney Urban Objects dataset for the proposed method. This data set contains a variety of common urban road objects scanned with a Velodyne HDL-64E LIDAR, collected in the CBD of Sydney, Australia. In the preprocessing step, the data set was loaded into memory by parsing the raw data file. After that, the scene properties were set such as background color, point color, etc. Then, the geometric centre of the point clouds was determined. YOLO did the classification and localization of the object in only one step which resulted in determining the position and category of the object directly at the output layer.

In the "Training" step, the mAP was around 70% which was rather low. However by using their voting mechanism they are able to achieve a classification accuracy of 98.42% on the Sydney Urban Objects data set. In the "Inference" step, an important parameter to tune was the threshold at which they considered the object to be classified. They performed an empirical study to gauge the most effective value and found that the optimal value was 58. Overall their method had higher accuracy than other existing methods.

Jannatul Mourey et al. focused on detecting fall detection for the elderly people using human detection algorithms [22]. They used OpenCV to implement the proposed algorithm. In addition to recognizing a fall, detection algorithm could analyze the severity of a fall by calculating how much time the subject took to rise to a stable condition if he/she was able to get up at all. The authors studied behavioral elements of the human body to detect the fall based on the sudden movement responses of the human body.

They selected Le2I data set to test their implementation. This data set is actually created to detect human falls in different environments and surroundings. The proposed method consisted of five stages: Video Analysis, Human Body Approximation, Fall Definition, Fall Detection, and Notification. For the video analysis part, they converted Le2I video to image frames. After that they subtracted background using two algorithm GMG and MOG2. After that they applied average filters to denoise the frames and carefully selected the Region of Interest (ROI) while the human body was approximated with a bounding box and an ellipse. Next, when the angle between the major axis and floor, and the size of the ellipse or the ratio of width to height of the bounding box exceeded the threshold, the event was considered a fall.

The model's accuracy was being tested using two methods. First one was using mAP and second one was using the annotation of the image frames. Proposed approach had the accuracy rate of 85% (accuracy method 1) and 86.6% (accuracy method 2).

Xianting Ke et al. proposed a race classification algorithm using iris image segmentation method [23]. The proposed method firstly made use of the merits of local Gabor binary pattern (LGBP) with support vector machine (SVM) and built an efficient classifier, LGBP-SVM, to partition iris images into the human eye and non-human eye images. Later, these two kinds of iris images are segmented by different strategies based on circular Hough transform with the active contour model. The LGBP-SVM extracted texture features in iris images and classifies these images into two groups such as human eye and non-human eye iris images. After that, the localizing region-based active contour model and circular Hough transform were separately employed to localize iris regions with different strategies. Their proposed method worked better to achieve a higher correct classification rate (CCR) and low equal error rate (EER). Their method mainly consisted of two methods – feature extraction and classification and the other one is iris image segmentation. Their first experiment was between human vs lion. The number of training samples was 8000 and the accuracy of the classification for LGBP-SVM was higher than 0.96 which was higher than the existing models like LGBP-KNN and LGBP-LDA. Lastly, they randomly selected 700 Asian and 700 White iris images for classification purpose. In here, LGBP-SVM also outperformed other models having the highest CCR of 99.92% with the EER of zero.

### C. Project Related Papers

Raman Maini et al. proposed an image enhancement technique based on two categories - Spatial Domain Methods and Frequency Domain Methods [7]. They used different transformations methods including Thresholding Transformations, Intensity Transformation, Logarithmic Transformations and Powers-Law Transformations and compared the results to show which transformation technique showed better outputs. For SDM, they dealt with the image pixels directly of the input images. To enhance the images they manipulated the pixel values. For FDM, they converted the images into frequency domain that means the Fourier Transform of the image was calculated. After that, all the image enhancement techniques were applied on the Fourier transform of the image and after that to get the resultant image Inverse Fourier Transform was applied. For the enhancement, they considered image brightness, contrast or the distribution of the gray levels. In this paper, authors only considered gray scaled images. They applied some point processing operation such as Thresholding Transformation, Intensity Transformation, Logarithmic Transformation, Powers-Law Transformation, Piecewise Linear Transformation and Gray level slicing. Before doing all these operation, they created negative version of the image.

The next part they did, they tried to process the histogram of the images rather their pixels value. They applied Histogram Equalization, Histogram Matching or Local Enhancement on the negative image. As concluding remarks, they discussed broadly about the outcome from each of the operation they applied on the negative image. Even though they explained each method in detail but they did not discuss about the computational cost. Also, they did not discuss on the point

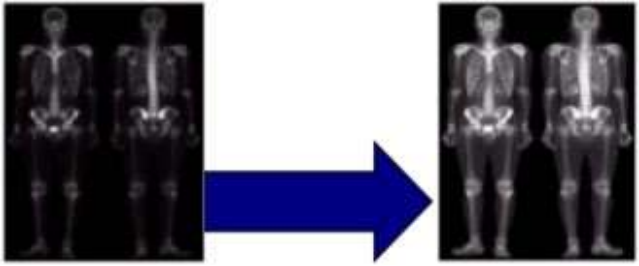

**Fig. 10: Enhanced image using SDM and FDM**

where these methods can be combined and show an efficient image enhancement.

Farzam Farbiz et al. showed a new fuzzy-logic-control based filter with the ability to remove the noise such as impulsive and Gaussian noise but they tried to preserve the edges and image details accordingly [25]. First they developed a fuzzy control filter (FCF) to preserve the edges but the filter could not perform well in terms of denoising Gaussian noise. After that authors modified the FCF to get the better output for smoothening Gaussian noise. All their filtering models were experimented with different image enhancement problems. In the FCF model, they took the general structure of a fuzzy if-then-else rule mechanism. Their approach is based on the idea that the algorithm will not let each point in the area of concern be uniformly fired by each of the basic fuzzy rules. They took the consideration of Mean-square-error (MSE) between the edges so that they can validate the edge preservation accuracy. They conducted the experiments based on many types of image enhancement problem such as edge preservation, impulsive noise, complex images in noise etc. And the FCF actually showed better performance than most of the existing algorithms. To get the better output for the Gaussian noise, they made two modifications. With the modified smoothing fuzzy control filter (SFCF), the authors showed that the SFCF worked better than the other filters for all noise levels. Also, SFCF has less computation time and is capable of image restoration in noise environments.

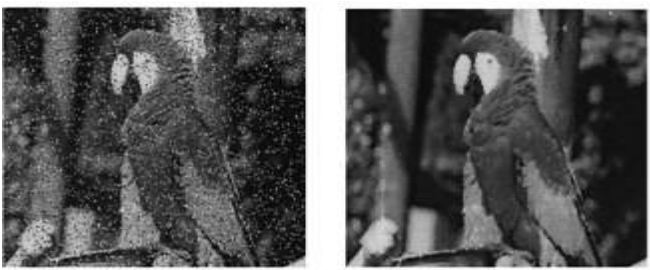

**Fig. 11: Result using fuzzy control filter (FCF)**

Kuldeep Singh et al. developed a robust contrast enhancement algorithm based on histogram equalization methods named Median-Mean Based Sub-Image-Clipped Histogram Equalization (MMSICHE) [24]. Their proposed algorithm has three steps. First they calculated the median and mean brightness values of the image pixels. After that the histogram is clipped using a plateau limit set as the median of the occupied intensity. And lastly, the clipped histogram is first bisected based on median intensity then further sub-divided into four sub images based on each sub images' mean intensity, also performing histogram equalization for each sub image. By this new approach they preserved the brightness of the images and also the image information content that means entropy. Subsequently they controlled over enhancement rate which is beneficial for electronics applications. They tried to work with the natural brightness enhancement of the image. Also the results showed that MMSICHE methods performed better than the other histogram equalization methods (WTHE, QDHE, RSWHE, SHMS, BHEPL and BHEPL-D) in terms of four quality factors such as average luminance, entropy, absolute mean brightness error (AMBE) and background gray level. In fig. 11, enhancement using different methods are being shown.

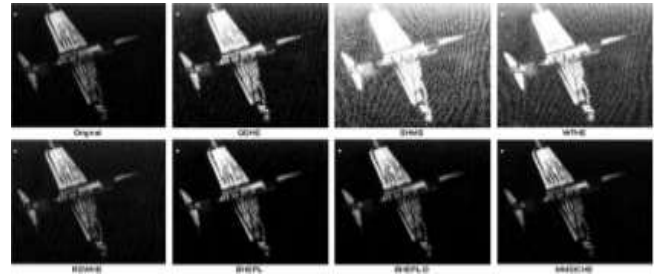

**Fig. 12: Enhancement results of U2 image**

The authors conducted the experiments using five test images – Tank, U2, Field, Copter and Hands. For Copter, Hands and Field image, the MMSICHE method produces the images having an average luminance which is actually relatively close to the original image. Likewise, in terms of AMBE, MMSICHE approach showed better performance than all other methods. Proposed algorithm has the least AMBE value for the three mentioned images.

In terms of the image information contents (entropy), MMSICHE has the highest entropy value which is almost equal to original image. Also, SHMS, WTHE and QDHE has amplified the noise but the proposed algorithm is free from any kind of noises. MMSICHE actually did produce images with contrast enhancement with close natural appearances.

Andrea Polesel et al. proposed an Adaptive Contrast Enhancement Algorithm that controls the contribution of the sharpening path in such a way that contrast enhancement happens in high detail areas and little or no image sharpening occurs in smooth areas [6]. The algorithm employs two directional filters whose coefficients are updated using a Gauss–Newton adaptation strategy. The algorithm introduced two directional filters whose coefficients were updated using a Gauss–Newton adaptation strategy. Using their approach, the result showed better image enhancement than other existing approaches.

S. Rajeshwari et al. concentrated on the efficient denoising and an improved enhancement technique of the medical images [5]. They worked with the average, median and wiener filtering for image denoising and an interpolation based Discrete Wavelet Transform (DWT) technique for resolution enhancement. Later, they compared the images using Peak Signal to Noise Ratio (PSNR).

Initially, authors took the MRI image as input and added some with Gaussian noise for experiment purpose. Among the Average, Median and Wiener filters, Wiener filter performed the deconvolution by inverse filtering that means by high pass filtering and also removed the noise with the low pass filtering. On the other hand, median filter could not perform well with large amount of Gaussian noise like the Average filter performed. Later they performed interpolation based discrete wavelet transform which actually did preserved the edges and the other information of the image. After that they compared the enhanced image with the denoised image in which enhanced image had better PSNR value.

Kuntal Kumar Pal et al. suggested, for image classification using Convolutional Neural Network, image preprocessing techniques play a vital role to increase the performance of the CNN [4]. Authors did take the CIFAR10 dataset and three different types of CNN for experiment purpose. They considered three different image preprocessing techniques such as Zero Component Analysis (ZCA), Mean Normalization and Standardization. Later they compared the results to show which preprocessing technique worked better for the above mentioned CNN architecture.

For all of the CNN, authors put the initial values of weights and biases of all the features using Normal Distribution. First Convolutional Network had 3 layers. A convolutional and pooling layer combined, a fully connected layer and a softmax layer with sigmoid function used as activation. Second Convolutional Network was similar to first CNN with one exception that is RELU was used as activation function. Third Convolutional Network had 4 layers. Two combined layers of convolutional and pooling, a fully connected layer and a softmax layer with ReLU used as activation function. The training and testing was done using GPU (GeForce820M) with python 2.7 and theano with cuda compilation tools (release 5.5, V5.5.0) on a machine having 8GBRAM and Intel core i3 processor. For the first CNN, ZCA outperformed Mean Normalization and Standardization having the accuracy of 64-68%. For the second CNN, ZCA outperformed having the accuracy of 67-69%. Also for the third CNN, ZCA did better job having the accuracy of 67-73%. They concluded as ZCA being the best image preprocessing technique for the image classification using CNN.

### III. Data Collection

PASCAL VOC 2007 is selecated as out training dataset [2]. The training data provided consists of a set of images; each image has an annotation file giving a bounding box and object class label. Classes including person, animal, vehicle, furniture, and so on. It is an ideal tool for object class recognition because it is standard and precise.

### IV. Proposed Method

Faster R-CNN [1] is an object detection architecture presented by Ross Girshick et al. in 2015. It is composed of 3 parts: Convolution layers to extract features, Region Proposal Network to judge the occurrence of an object and circling its bounding, Classes and Bounding Boxes prediction which is fully connected neural networks predict object class and bounding boxes. As a standard network for object detection, it is famous and widely used.

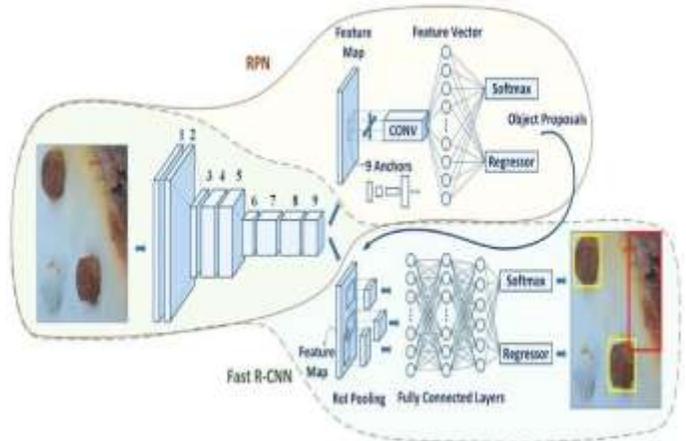

**Fig. 13: Faster R-CNN network**

YOLO (You Only Look Once) [8] is one of the most powerful real time object detection algorithm.It only needs the image to pass one time through its network.This ensures that Yolo can complete object recognition quickly.

Firstly proposed by M.A. Hearst, A support vector machine (SVM)[9] is a supervised machine learning model that used in classification problems. SVM model will try to find a hyperplane (as shown in the figure below) after giving the model sets of labeled training data. With this hyperplane, SVM can classify new data into existing categories. SVM is also a fundamental model for machine learning with wide usage.

We separate our work into two stages. In the first stage, we look into the statistical data of PASCAL VOC [2] and use histograms to study the statistical characteristics of the data, such as the distribution of resolution and information on the color of the image data. Based on this information and related work, we will use image processing to adjust the dataset. Then we feed the adjusted data into a fixed faster R-CNN model. From the model's training time and performance, we will have an intuition about what image processing methods are promising choices for object detection. In the next stage, our job is adjusting the parameters of possible image processing methods using Machine learning. One possible way is that we design a Neural network for a picked subset of PASCAL VOC [2]. This subset is designed to approximately have the same statistical information as the full dataset.

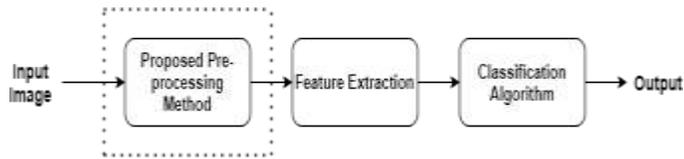

Fig. 14: Diagram for the proposed pre-processing method

For the second stage, we implement models such as Faster R-CNN [1] or SVM [9] on the sub-dataset to learn to achieve the highest accuracy in a given time or to use the least time for a fixed accuracy. The reason to use the subset is that image processing methods like may take a lot of time to finish. But when the dataset is small, we can afford the processing time. The outcome of this procedure will give us intuition about what FR-CNN 'likes' to see. Besides, it can provide us the initial parameter for full dataset training. At last, we apply the outcome to all images in the dataset then evaluate the model's efficiency and accuracy. Our expectation is getting a faster convergence FR-CNN model with similar or even better accuracy.

## V. Individual Responsibility

Below table indicates the team members' overall participation on this project -

| Team Members' Participation ||
|---|---|
| **Tasks** | **Team Members** |
| Review of Related Work | Tanzil/Huyue |
| Analysis of the Data | Huyue |
| Analysis of Current Method Scope and Feasibility | Tanzil |
| Development of Implementation Pipeline | Huyue |
| Data Preprocessing | Tanzil |
| Implementation of Proposed Method | Tanzil/Huyue |
| Report Writing | Tanzil/Huyue |